\documentclass{article}

    \PassOptionsToPackage{numbers, compress}{natbib}

 \usepackage[preprint]{neurips_2026}

\usepackage[utf8]{inputenc} %
\usepackage[T1]{fontenc}    %
\usepackage{hyperref}       %
\usepackage{url}            %
\usepackage{booktabs}       %
\usepackage{amsfonts}       %
\usepackage{nicefrac}       %
\usepackage{microtype}      %
\usepackage{xcolor}         %
\usepackage{graphicx}
\usepackage{amsmath}
\usepackage{multirow}
\usepackage{wrapfig}

\title{DARE: Dual-Level Adversarial Learning with Domain-Aware Regularization for Whole Slide Image Classification}

\author{%
  \textbf{Yuxuan Jiang}$^{1,2}$\thanks{Equal contribution.}
  \qquad
  \textbf{Haichuan Dong}$^{1,2}$\footnotemark[1]
  \qquad
  \textbf{Manning Wang}$^{1,2}$\thanks{Corresponding author.}
  \\[0.5em]
  $^{1}$ Digital Medical Research Center, School of Basic Medical Sciences, Fudan University, China \\
  $^{2}$ Shanghai Key Laboratory of Medical Image Computing and \\
  Computer Assisted Intervention, China \\
  \texttt{\{25211010025,25111010027\}@m.fudan.edu.cn}, \;\texttt{mnwang@fudan.edu.cn}
}

\begin{document}

\maketitle

\begin{abstract}
Pathological whole slide images (WSIs) exhibit substantial domain shifts across hospitals due to variations in scanning devices, staining protocols, and tissue preparation procedures, leading to significant performance degradation when models are applied to data from different sources. Although unsupervised domain adaptation (UDA) has shown strong performance in transferring models from a labeled source domain to an unlabeled target domain in natural images, it remains relatively underexplored in computational pathology. The ultra-high resolution of WSIs and their pronounced inter-slide and intra-slide heterogeneity pose unique challenges for domain adaptation, often rendering existing UDA methods unstable and susceptible to negative transfer. To address these challenges, we propose Dual-level Adversarial learning with domain-aware REgularization (DARE), a plug-and-play UDA framework for WSI classification. DARE incorporates adaptive pseudo-labeling within a teacher-student architecture to provide stable supervision for the target domain, alongside adversarial alignment of both patch-level and slide-level embeddings. Together, these components enable the joint optimization of semantic representations at both levels for cross-domain alignment. We further introduce domain-aware attention masking to regularize attention and improve cross-domain generalization. Extensive experiments on four public WSI datasets demonstrate that DARE consistently outperforms state-of-the-art approaches across various domain shifts.

\end{abstract}

\section{Introduction}
\begin{figure}[t]
  \centering
  \includegraphics[width=\linewidth]{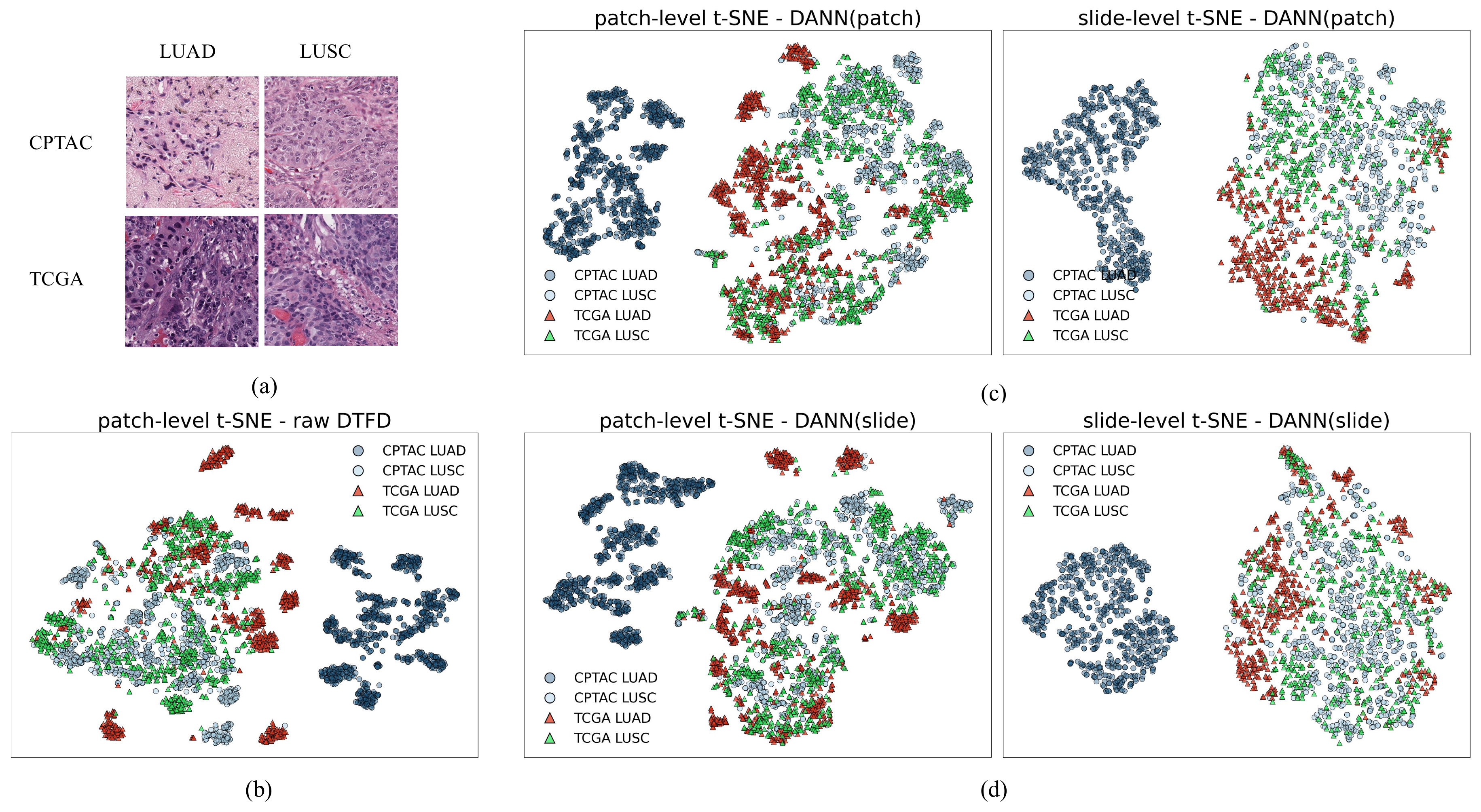}
  \caption{Illustration of domain discrepancy and DANN-based feature alignment on the CPTAC \(\rightarrow\) TCGA (NSCLC) task. (a) Example patches. (b) t-SNE of original DTFD-aggregated slide-level features. (c) Applying DANN to patch-level features: (left) patch-level, (right) slide-level t-SNE. (d) Applying DANN to slide-level features: (left) patch-level, (right) slide-level t-SNE. CONCH is used for feature extraction.}
  \label{fig:intro}
\end{figure}

Classification of pathological whole slide images (WSIs) plays an important role in clinical cancer detection, subtyping, and treatment response prediction, among others.
Recently, deep learning has achieved remarkable success in WSI classification tasks, reaching or exceeding human-level performance~\cite{cheplygina2019not,qu2022towards,srinidhi2021deep,rony2019deep}.
However, the performance of these models often degrades when deployed across different clinical centers~\cite{boutaj2025controllable}. This degradation is mainly caused by domain shifts between medical institutions, including variations in scanning devices, staining protocols, and tissue preparation procedures. 
As shown in Fig.~\ref{fig:intro}a, noticeable staining differences can be observed between the CPTAC-NSCLC and TCGA-NSCLC datasets.

Due to the extremely high resolution of WSIs, reaching up to 100,000 \(\times\) 100,000 pixels, they are typically divided into a large number of patches and modeled under a multiple instance learning (MIL) framework~\cite{lu2021ai,lu2022federated,mahmood2019deep,nagpal2019development}, where a patch-level feature extractor is applied first to obtain local patch representations and an attention-based aggregator is then used to aggregate these patch features into a slide-level feature for classification under slide-level supervision~\cite{li2022comprehensive,salvi2021impact,wang2018revisiting}.
In recent years, pathology foundation models have advanced rapidly, and the current mainstream approach is to use a frozen foundation model encoder for patch-level feature extraction. Although foundation models are generally considered more generalizable, they still cannot fully address the domain shift problem. We extract patch-level features from the CPTAC-NSCLC and TCGA-NSCLC datasets using CONCH~\cite{lu2024avisionlanguage} as the encoder and visualize their feature distributions in Fig.~\ref{fig:intro}b. Substantial distribution shifts can be observed across the two datasets in the feature space. Such distribution shifts make it still difficult for models trained on one domain to generalize to another domain. This requires additional annotations from pathologists on the target domain data, followed by retraining on the labeled target data, which significantly increases both time and labor costs.

\begin{wrapfigure}{r}{0.35\linewidth}
  \centering
  \includegraphics[width=\linewidth]{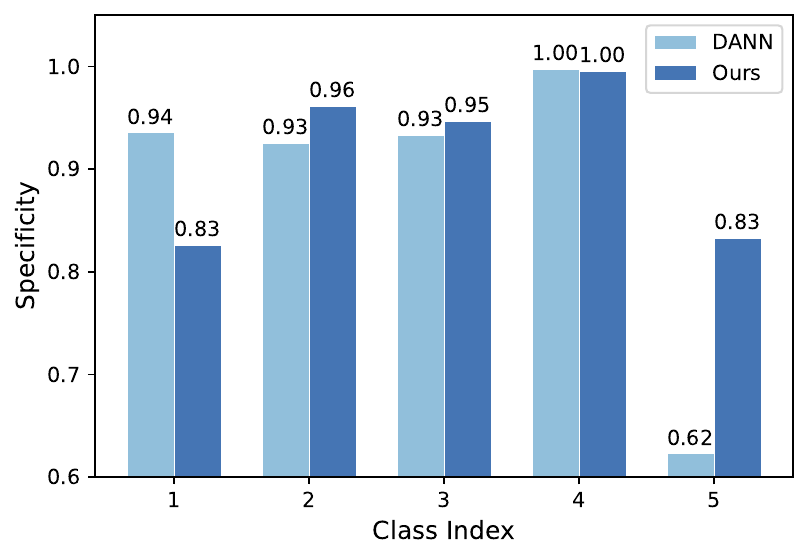}
  \caption{Specificity comparison on IPD \(\rightarrow\) TCGA (GBMLGG).}
  \label{fig:specificity}
\end{wrapfigure}

Unsupervised domain adaptation (UDA) offers a promising paradigm to address this issue. It enables a model trained on a labeled source domain to transfer to an unlabeled target domain~\cite{ben2010theory,saenko2010adapting}. Many UDA methods have been developed for natural images, but they usually do not generalize well to WSI classification. Our experiments show that directly applying these methods to pathology often leads to instability or even negative transfer. There are some UDA studies in the computational pathology field, and they mainly operate at the image level. They focus on color normalization~\cite{macenko2009method,bejnordi2015stain,tellez2019quantifying} or style transfer~\cite{shaban2019staingan,wagner2021structure}. However, these methods fail to capture high-level semantic structures in the feature space. Moreover, these image-level methods are not well aligned with the current paradigm of patch-level feature extraction using foundation models.

To improve UDA performance for WSI classification under the current mainstream pipeline of extracting patch features using a frozen pathology foundation model, we design a novel framework to perform domain adversarial learning at the feature level. The framework is motivated by the specific characteristics of UDA in WSI classification. First, as shown in Fig.~\ref{fig:intro}c and Fig.~\ref{fig:intro}d, we apply the classical domain adversarial learning method DANN~\cite{ganin2016domain} at the patch level and at the slide level, respectively, and visualize the resulting feature distributions. We observe that applying DANN at a single granularity yields poor feature separation at both levels: neither patch-level features nor slide-level features can be well separated by subtype in the target domain. WSIs naturally exhibit a two-level structure, where patch-level and slide-level features capture different granularities of information. Domain shifts can exist at both levels, so aligning only one may not be enough.
Second, most regions within a WSI correspond to background or normal tissue, while diagnostically relevant regions are often sparse, and when aggregating patch features using the attention mechanism, the final prediction highly depends on a small subset of discriminative patches~\cite{ilse2018attention,shi2020loss,keshvarikhojasteh2024multi,xiang2023multi}. The bias amplification effect in the attention mechanisms for aggregating patch-level features can harm generalization ~\cite{lu2021data,zhang2024attention}. 
In cross‑domain settings, a domain‑agnostic aggregator cannot distinguish whether a highly attended patch carries transferable semantics or domain‑specific shortcuts. 
This calls for a domain-aware attention regularization strategy that can suppress domain-specific easy patches and encourage learning from harder, domain-invariant evidence.
Third, we compare the per-class specificity of our method and slide-level DANN. As illustrated in Fig.~\ref{fig:specificity}, DANN suffers from poor specificity on the fifth class, revealing that global domain alignment does not guarantee class-wise alignment. Achieving category-level alignment across domains requires a mechanism that provides class-specific supervision in the target domain to separate the distributions of different categories.

Accordingly, we propose Dual-level Adversarial learning with domain-aware REgularization (DARE), a framework for WSI domain adaptation that integrates dual-level adversarial alignment, domain-aware instance selection, and adaptive pseudo-label supervision. Specifically, DARE performs patch-level adversarial learning to reduce local feature discrepancies between the source and target domains. Meanwhile, patch-level domain predictions are combined with attention scores to identify and mask highly attended target patches that exhibit strong source-domain characteristics. The retained patches are then re-aggregated into target slide representations and adversarially aligned with source slide representations. In this way, patch-level domain information guides local instance selection and determines which tissue evidence contributes to slide-level alignment, coupling adaptation across the two levels. Furthermore, adaptive soft pseudo-label supervision from the unmasked teacher guides the masked student to preserve class semantics while learning from the remaining evidence. Together, these components connect local feature alignment, domain-aware instance selection, global distribution alignment, and semantic preservation.

Our main contributions are summarized as follows:

\begin{itemize}
\item We propose a domain-aware attention masking strategy that combines patch importance with domain predictions to identify and suppress highly attended, source-specific patches. This strategy incorporates
domain information into instance selection and encourages the model to learn from more transferable tissue evidence.

\item We develop a cross-level adaptation mechanism in which patch-level
domain predictions guide instance selection, and the retained instances
are re-aggregated for slide-level adversarial alignment. Adaptive
teacher-student pseudo-label supervision preserves class semantics
during this process, coupling local instance selection with global
domain adaptation.

\item We evaluate DARE on four public datasets and demonstrate superior
performance over state-of-the-art UDA methods, including under limited
target-domain data. Experiments with alternative attention-based MIL
backbones and pathology foundation encoders further demonstrate its
plug-and-play applicability and compatibility with different feature
representations.
\end{itemize}

\section{Related Work}

\subsection{Multiple Instance Learning for WSI Classification}
MIL for WSI classification follows a dual-level structure: patch-level features are first extracted from individual patches, and then aggregated into slide-level representations for final prediction.
Most existing MIL methods are based on attention mechanisms for feature aggregation. For example, ABMIL~\cite{ilse2018attention} assigns an attention score to each patch and performs weighted aggregation to obtain slide-level features. 
DTFD-MIL~\cite{zhang2022dtfd} further improves MIL from the perspective of feature distribution modeling.
These methods typically rely on pretrained ResNet~\cite{chen2020simple} models for patch-level feature extraction. With the emergence of pathology foundation models, their use as feature extractors for patch-level representations has become a dominant paradigm in WSI analysis. For example, HistAUG~\cite{boutaj2025controllable} uses CONCH~\cite{lu2024avisionlanguage} and UNI~\cite{chen2024uni} for feature extraction and introduces spatial augmentation on top of attention-based frameworks to improve generalization.
MAPLE~\cite{zhou2025maple} uses PLIP~\cite{huang2023visual} as the patch encoder and proposes an entity-guided cross-attention mechanism.

Although these methods have achieved significant progress in single domain settings, most of them assume that training and test data follow the same distribution~\cite{zhu2024dgr,zhang2026rethinking}. They largely lack explicit modeling of cross-center distribution shifts, which limits their generalization in real-world clinical scenarios.

\subsection{Unsupervised Domain Adaptation}
Unsupervised domain adaptation aims to learn models that generalize well to an unlabeled target domain using labeled source domain data and unlabeled target domain data. Existing methods can be broadly categorized into two groups: distribution alignment methods and self-training methods.
Distribution alignment methods aim to reduce the discrepancy between source and target feature distributions. They can be further divided into two subcategories: methods based on statistical distance and methods based on adversarial learning.
Statistical distance methods minimize explicit divergence measures, such as maximum mean discrepancy or optimal transport, to align the distributions~\cite{long2015learning,sun2016deep,zellinger2019robust,courty2016optimal,morerio2017minimal}. 
Adversarial learning methods introduce a domain classifier and train the feature extractor in an adversarial manner to produce domain-invariant representations~\cite{ganin2016domain,long2018conditional,saito2018maximum,tzeng2017adversarial,chadha2019improved}. 
In computational pathology, RAM-MIL~\cite{cui2023retrieval} is based on statistical distance and leverages optimal transport for nearest-neighbor retrieval to reduce slide-level domain shifts. DETA~\cite{shou2025graph} is based on adversarial learning and employs an adversarial perturbation strategy to augment source domain features, performing slide-level adversarial alignment.
However, these distribution alignment methods largely ignore the hierarchical patch-slide relationship inherent to MIL-based WSI analysis.
In contrast, self-training methods generate pseudo-labels for target domain samples and iteratively refine the model using these pseudo-labels to improve target domain discrimination~\cite{kim2019self,liu2021cycle,kumar2020understanding,gu2020spherical}.
Therefore, our method DARE integrates dual-level adversarial learning with self-training, explicitly aligning features at both patch and slide levels while introducing pseudo-labeling to provide supervision for the target domain.

\section{Method}
\subsection{MIL Problem Formulation}

For the source domain, we denote the dataset as 
\(S = \{(I_i^s, Y_i^s)\}_{i=1}^{N_s}\), where each WSI \(I_i^s = \{p_{i,j}^s\}_{j=1}^{n_i}\) is composed of \(n_i\) patches. 
The label \(Y_i^s \in \{1,\dots,C\}\) denotes the slide-level annotation.
For the target domain, we have an unlabeled dataset 
\(T = \{I_i^t\}_{i=1}^{N_t}\), which contains only WSIs without annotations.
In attention-based MIL, a feature extractor \(f(\cdot)\) is first applied to each patch to obtain patch-level features \(h_{i,j}^D\), where \(D \in \{s,t\}\). 
An attention-based aggregator \(\mathrm{Agg}(\cdot)\) is then used to obtain slide-level features \(F_{i}^D\) by pooling patch-level features.
Finally, a classifier \(g(\cdot)\) operates on the slide features to produce the predictions 
\(\hat{y}_i^D = g(F_{i}^D)\).
Under the unsupervised domain adaptation setting, the model is trained using only slide-level labels from the source domain while leveraging unlabeled data from the target domain to learn representations with strong cross-domain generalization.

\subsection{Method Overview}
\begin{figure}[htbp] 
  \centering
  \includegraphics[width=\linewidth]{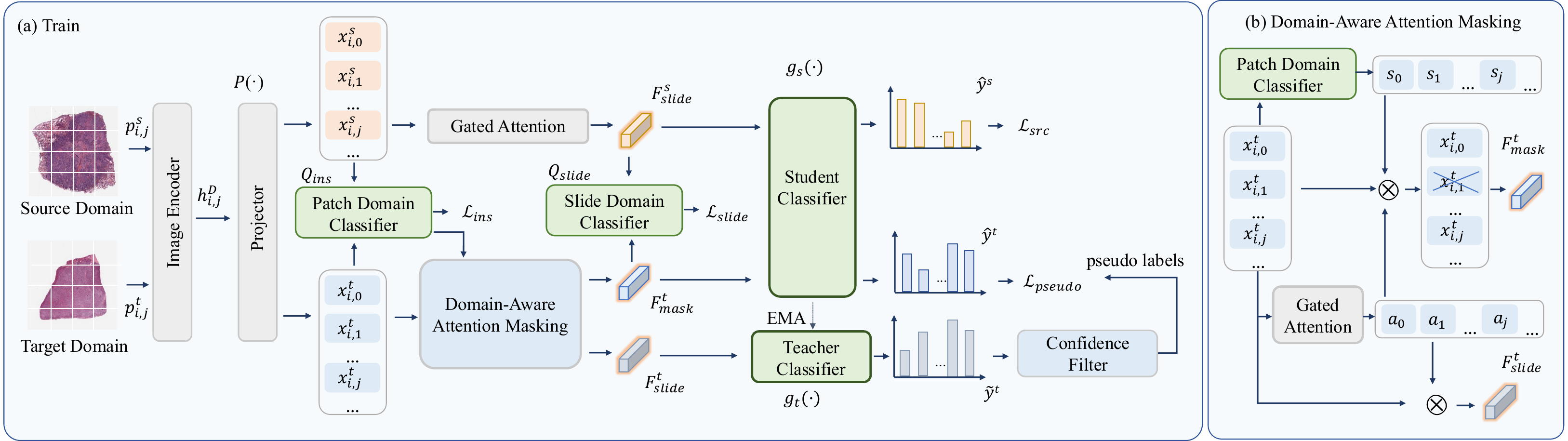}
  \caption{Overview of the DARE framework. (a) Training pipeline with multi-granularity domain alignment, attention masking, and pseudo-label learning. (b) Domain-aware attention masking: masking patches guided by domain and attention distributions.}
  \label{fig:method}
\end{figure}

As shown in Fig.~\ref{fig:method}, the WSI is first divided into patches, and a pathology foundation model is used to extract patch-level features \(h_{i,j}^D\). After a projection network \(P(\cdot)\), these features are mapped into an embedding space, yielding \(x_{i,j}^D\). Patch-level embeddings \(x_{i,j}^D\) are aggregated by an attention network to obtain slide-level embeddings for the source and target domains, denoted as \(F_{slide}^{D}\). For the source domain data, we follow the standard WSI classification pipeline. The source slide-level embedding \(F_{slide}^{s}\) is fed into a classifier shared with the student network to give a slide-level prediction \(\hat{y}^s\).

In the proposed DARE framework, we first adopt a patch-level domain classifier \(Q_{ins}\) to adversarially align patch-level embeddings of the source and target domains. Meanwhile, for each target domain patch, \(Q_{ins}\) outputs a probability that the patch belongs to the source domain, and the attention module produces an attention score. By jointly considering these two quantities, we select and mask the Top-\(K\) patches. The remaining patch embeddings are re-aggregated to obtain a masked slide-level embedding \(F_{mask}^t\) for the target domain.
We further adopt a slide-level domain classifier \(Q_{slide}\) to adversarially align \(F_{slide}^{s}\) from the source domain and \(F_{mask}^t\) from the target domain.
In parallel, we can also aggregate the target domain patch embeddings through standard attention without masking, producing an unmasked slide-level embedding \(F_{slide}^{t}\).
Finally, a teacher-student pseudo-labeling framework is adopted to provide supervisory signals for the target domain. The teacher classifier takes the unmasked embedding \(F_{slide}^{t}\) as input and generates pseudo-labels. The student classifier takes the masked embedding \(F_{mask}^t\) and learns to predict these pseudo-labels. Importantly, the source and target domains share the same student classifier: source slide embeddings \(F_{slide}^{s}\) are supervised by ground-truth labels, while masked target embeddings \(F_{mask}^t\) are supervised by pseudo-labels from the teacher.
In inference, both domain classifiers and the attention masking module are removed, leaving only the attention-based MIL aggregator and the student classifier for prediction.

\subsection{Dual-Level Domain Adversarial Learning}
The domain shift between the source and target domains is reflected not only in local cellular and tissue-level texture appearance at the patch level, but also in global contextual distributions of the tissue microenvironment at the slide level. To address this, we introduce a dual-level domain adversarial module.

Specifically, the patch-level domain classifier \(Q_{ins}\) distinguishes whether a patch embedding \(x^{D}_{i,j}\) comes from the source or the target domain. Its adversarial loss \(\mathcal{L}_{ins}\) is defined as:

\begin{equation}
\mathcal{L}_{ins}=\mathrm{CE}(Q_{ins}(x^{D}_{i,j}),d),
\end{equation}

\(\mathrm{CE}(\cdot)\) denotes the standard cross-entropy function, and \(d\in\{0,1\}\) is the domain label. To achieve cross-domain embedding alignment, we introduce an adversarial learning scheme between the projector \(P(\cdot)\) and the domain classifier \(Q_{ins}\). The projector \(P(\cdot)\) is encouraged to produce domain-invariant representations. Formally, this process can be cast as a mini-max optimization problem: the domain classifier \(Q_{ins}\) aims to maximize domain classification accuracy, while \(P(\cdot)\) aims to minimize this ability, thereby aligning the distributions of patch embeddings from the source and target domains.
In practice, we insert a gradient reversal layer (GRL) after the patch embeddings. During the forward pass, the GRL acts as an identity mapping and passes embeddings to the domain classifier. During backpropagation, the GRL multiplies the gradients from the domain classifier by a negative coefficient, effectively reversing their direction before passing them to the projector. As a result, the projector is optimized to confuse the domain classifier, leading to more domain-invariant embeddings.

Then, we introduce a slide-level domain classifier \(Q_{slide}\) for slide-level embeddings and enforce adversarial alignment. For the target domain, we apply the domain-aware attention masking mechanism to patch embeddings to obtain an aggregated slide embedding \(F_{mask}^t\). We align it with the source slide embedding \(F_{slide}^{s}\) by minimizing the slide-level adversarial loss \(\mathcal{L}_{slide}\).

\subsection{Domain-Aware Attention Masking}
We revisit the role of attention under cross-domain settings and introduce a domain-aware attention masking strategy. 
Specifically, patches with both high attention scores and high source-domain probability from \(Q_{ins}\) are treated as domain-specific easy patches and are dynamically masked. 

As illustrated in Fig.~\ref{fig:method}b, for a target domain WSI indexed by \(i\), let \(X^{t}_{i} = \{x^{t}_{i,1}, x^{t}_{i,2}, \dots, x^{t}_{i,n_i}\}\) denote its patch-level embedding set, where \(n_i\) is the number of patches. 
A gated attention mechanism is applied to compute an attention score for each patch. The attention weights \(a_{i,j} \in [0,1]\) and the aggregated slide-level embedding \(F_{slide}^{t}\) are defined as:

\begin{equation}
    a_{i,j} = \frac{\exp \left( w^{T} \big( \tanh(V x^{t}_{i,j}) \odot \sigma(U x^{t}_{i,j}) \big) \right)}
{\sum_{q=1}^{n_i} \exp \left( w^{T} \big( \tanh(V x^{t}_{i,q}) \odot \sigma(U x^{t}_{i,q}) \big) \right)},
\end{equation}
\begin{equation}
    F_{slide}^{t} = \sum_{j=1}^{n_i} a_{i,j} \, x^{t}_{i,j},
\end{equation}

where \(w\), \(V\), and \(U\) are learnable parameters of the attention network, and \(\odot\) denotes element-wise multiplication in the gating mechanism.

Meanwhile, for each target domain patch \(x^{t}_{i,j}\), we use a patch-level domain classifier \(Q_{ins}\) to obtain a probability \(p_{i,j}\) that it belongs to the source domain.
We define a joint masking score by combining the attention weight and the domain probability as \(m_{i,j} = p_{i,j} \times a_{i,j}\).
We rank all patches in \(X^{t}_{i}\) by \(m_{i,j}\) and mask the Top-\(K\) ones, where \(K = \lceil r \cdot n_i \rceil\) and \(r\) is a predefined masking ratio.
This yields the masked embedding set \(\widetilde{X}^{t}_{i}\).
The attention weights over the remaining patches in \(\widetilde{X}^{t}_{i}\) are re-normalized to sum to one and aggregated into the masked slide-level embedding \(F_{mask}^t\).

\subsection{Teacher-Student Based Adaptive Pseudo-label Learning}
For each target domain slide, the domain-aware attention masking module produces an unmasked slide embedding \(F_{slide}^{t}\) via standard attention aggregation, and a masked slide embedding \(F_{mask}^t\) by suppressing domain-specific easy patches and re-aggregating the remaining ones.
\(F_{mask}^t\) flows into the student branch, passes through a student classifier \(g_s(\cdot)\) shared with the source domain, and produces a prediction \(\hat{y}^t\).
\(F_{slide}^{t}\) flows into the teacher branch and passes through a teacher classifier \(g_t(\cdot)\) to generate soft pseudo-labels \(\tilde{y}^t\).
The student is trained to match these pseudo-labels.
The teacher receives no gradient updates; its parameters are an exponential moving average (EMA) of the student parameters.

To improve pseudo-label quality and training stability, we further adopt pseudo-label memory maintenance for building a global bank with ensemble predictions, adaptive threshold selection for filtering noisy labels via smoothly evolving per-class thresholds, and class-balanced weighting for mitigating class imbalance, as detailed below.

\textbf{Pseudo-Label Memory Maintenance.}
We maintain a global pseudo-label bank that stores soft pseudo-labels for all target domain slides.
At the beginning of each epoch, we use the teacher network's attention parameters from the end of the previous epoch to aggregate the patch features of every target domain slide, producing their unmasked slide embeddings \(F_{slide}^{t}\). 
The teacher classifier then processes these embeddings to produce soft pseudo-labels, which are stored in the bank.
The soft pseudo-label for each target slide is computed as:

\begin{equation}
\tilde{y}^t={Softmax}(g_t(F_{slide}^{t})),
\end{equation}
where \(g_t\) denotes the teacher network's classifier.

\textbf{Adaptive Threshold Selection.}
From the pseudo-label bank, we can obtain a soft pseudo-label \(\tilde{y}_i\) and its confidence \(\mu_i\) for every target sample.
To filter out noisy pseudo-labels, we adopt a per-class adaptive threshold.
For each class \(c\), we collect the confidences of all samples predicted as class \(c\) into a set \(\mu^c\) and compute its median as the initial threshold \(\hat{\tau}_c\).
To prevent abrupt fluctuations, we smooth \(\hat{\tau}_c\) via an EMA and clip the update to a maximum change \(\Delta\) from the previous threshold, resulting in the current threshold \(\tau_c^{(t)}\).
\begin{equation}
\tilde{\tau}_c = \alpha\cdot\tau_c^{(t-1)} + (1-\alpha)\cdot\hat{\tau}_c,
\end{equation}
\begin{equation}
\tau_c^{(t)} = \varphi\bigl(\tilde{\tau}_c,\ \max(\tau_{\min},\ \tau_c^{(t-1)}-\Delta),\ \min(\tau_{\max},\ \tau_c^{(t-1)}+\Delta)\bigr),
\end{equation}
where \(\varphi\) denotes the clipping function, \(\alpha\) is the smoothing coefficient, \(\Delta\) is a hyperparameter, and \([\tau_{\min},\tau_{\max}]\) prevents excessively low or high extreme threshold values. Based on the updated \(\tau_c^{(t)}\), we select the sample set \(A_c\) whose confidences satisfy \(\mu_i \ge \tau_c^{(t)}\).
The selected set for class \(c\) is \(A_c = \{ i \mid \mu_i \ge \tau_c^{(t)} \}\), which is then passed to the class-balanced weighting and pseudo-label loss.

\textbf{Class-Balanced Weighting of Pseudo Samples.}
After completing the selection for all classes, to account for potential class imbalance, we compute a class-specific normalized distillation weight \(w_c\) for class \(c\) based on the dynamic sample counts. The weight is inversely proportional to the number of accepted samples \(|A_c|\) for that class and is normalized by the total number of classes \(C\):

\begin{equation}
w_c = \frac{C\cdot |A_c|^{-1}}{\sum_{j=1}^{C} |A_j|^{-1}}.
\end{equation}

Let \(A = \bigcup_{c=1}^{C} A_c\) be the set of all accepted target samples, and \(c(i)\) be the pseudo-label class of sample \(i\). Under this mechanism, hard classes with fewer samples receive higher weights in the pseudo-supervision loss. The student network then aligns its predictions to the teacher-generated soft labels using KL divergence:
\begin{equation}
\hat{y}_i^t = {Softmax}\!\left(g_s(F_{mask,i}^t)\right),
\end{equation}
\begin{equation}
\mathcal{L}_{pseudo}=\frac{1}{|A|}\sum_{i\in A} w_{c(i)}\cdot \mathrm{KL}\bigl(\tilde{y}_i^t \parallel \hat{y}_i^t\bigr),
\end{equation}
where \(g_s\) is the student network's classifier.

The final total training loss is:
\begin{equation}
\mathcal{L}=\lambda_{ins}\mathcal{L}_{ins}+\lambda_{slide}\mathcal{L}_{slide}+\lambda_{pseudo}\mathcal{L}_{pseudo}+\mathcal{L}_{src},
\end{equation}
where \(\mathcal{L}_{src}\) denotes the supervised cross-entropy loss on the source domain.

\section{Experiments}
\label{sec:Evaluation Metrics}

\subsection{Experimental Setup}
\textbf{Datasets.} We evaluate our method on four public datasets, including TCGA-NSCLC, CPTAC-NSCLC, TCGA-GBMLGG, and IPD-Brain, and compare it with state-of-the-art approaches. TCGA-NSCLC and CPTAC-NSCLC are used for binary classification of Lung Adenocarcinoma (LUAD) versus Lung Squamous Cell Carcinoma (LUSC), while TCGA-GBMLGG and IPD-Brain are used for five-class classification of glioma subtypes. These datasets differ in both task settings and data characteristics, and exhibit substantial domain shifts in staining styles and patient populations. This setting provides a challenging benchmark for assessing cross-domain generalization.

\textbf{Implementation and Evaluation Details.}
We adopt the data preprocessing pipeline of CLAM~\cite{lu2021data}. WSIs are partitioned into patches of size \(512 \times 512\). Features for each patch are extracted in advance using the CONCH~\cite{lu2024avisionlanguage} encoder, resulting in 512-dimensional representations. Our method and the natural image based baselines are implemented within the DTFD-MIL~\cite{zhang2022dtfd} framework. Both the patch feature aggregators used in the source and target branches and the classifiers are based on the DTFD architecture.
We report ACC, F1-score, and AUC in the main experiments, while the ablation studies use AUC as the primary metric.
The source domain is divided into training and validation sets with a ratio of 8:2. For the target domain, we perform 5 random splits. In each split, the target domain data is divided into training and test sets with a ratio of 5:5.

\subsection{Main Results}

We report the performance of our method and competing approaches across different datasets in Table~\ref{tab:main_resluts}. Among the baselines, we include commonly used pathology models, DTFD~\cite{zhang2022dtfd} and CLAM~\cite{lu2021data}, which are trained on the source domain and directly evaluated on the target domain. We also compare with recent domain adaptation methods specifically designed for pathology images, namely DETA~\cite{shou2025graph}, RAM-MIL~\cite{cui2023retrieval}, and HASD~\cite{liu2025hasd}. However, given the limited number of domain adaptation methods in computational pathology, we further adapt a range of representative approaches from natural image analysis to the pathology setting, including MMD~\cite{long2015learning}, DANN~\cite{ganin2016domain}, CDAN~\cite{long2018conditional}, MCD~\cite{saito2018maximum}, SRDC~\cite{tang2020unsupervised}, CORAL~\cite{sun2016deep}, DRL~\cite{kim2026distributionally}, and SimProF~\cite{wang2025simprof}. All methods are compared comprehensively with our proposed approach.

The original DRL method relies on target domain labels for hyperparameter tuning, which are not available in the UDA setting. Therefore, we report DRL results both without and with access to target domain labels for a fair comparison. Our method achieves the best performance on most metrics across the four adaptation settings, even when compared with DRL tuned using target domain labels. On CPTAC \(\rightarrow\) TCGA (NSCLC), our method improves AUC by 8.98 points over the second-best method.
Meanwhile, we observe that many methods exhibit clear negative transfer compared with direct inference without adaptation. In particular, methods originally designed for natural image domain adaptation do not generalize well to pathological images. 
Notably, on the challenging CPTAC \(\rightarrow\) TCGA setting, several compared methods collapse to predicting a single class, resulting in the same ACC and F1. We provide a detailed analysis of this phenomenon in Appendix~\ref{sec:mode_collapse_analysis}.

\begin{table}
  \centering
  \caption{Experimental results across four datasets. For each setting, one dataset is used as the source domain and another as the target domain. The best and second-best results are highlighted in red and blue, respectively. Methods with the suffix “-P” are applied at the patch-level embedding space, while other adapted methods operate on slide-level embeddings by default.}
  \label{tab:main_resluts}
  \setlength{\tabcolsep}{1.25pt}
  \resizebox{\textwidth}{!}{
    \begin{tabular}{*{13}{c}}
      \toprule
      \multirow{2}{*}{methods} & \multicolumn{3}{c}{TCGA \(\rightarrow\) IPD (GBMLGG)} & \multicolumn{3}{c}{IPD \(\rightarrow\) TCGA (GBMLGG)} & \multicolumn{3}{c}{CPTAC \(\rightarrow\) TCGA (NSCLC)} & \multicolumn{3}{c}{TCGA \(\rightarrow\) CPTAC (NSCLC)} \\
      \cmidrule(lr){2-4}\cmidrule(lr){5-7}\cmidrule(lr){8-10}\cmidrule(lr){11-13}
      & AUC & ACC & F1 & AUC & ACC & F1 & AUC & ACC & F1 & AUC & ACC & F1 \\
      \midrule
      DTFD~\cite{zhang2022dtfd} & \(\,73.74{\scriptscriptstyle \pm0.58}\) & \(\,53.66{\scriptscriptstyle \pm0.64}\) & \(\,25.18{\scriptscriptstyle \pm3.18}\) & \(\,80.48{\scriptscriptstyle \pm2.41}\) & \(\,43.00{\scriptscriptstyle \pm2.04}\) & \(\,29.07{\scriptscriptstyle \pm1.61}\) & \(\,57.46{\scriptscriptstyle \pm1.30}\) & \(\,50.22{\scriptscriptstyle \pm2.03}\) & \(\,66.85{\scriptscriptstyle \pm1.80}\) & \(\,87.43{\scriptscriptstyle \pm2.93}\) & \(\,54.24{\scriptscriptstyle \pm1.44}\) & \(\,65.69{\scriptscriptstyle \pm1.22}\) \\
      CLAM~\cite{lu2021data} & \(\,80.85{\scriptscriptstyle \pm0.74}\) & \(\,61.52{\scriptscriptstyle \pm1.21}\) & \(\,32.43{\scriptscriptstyle \pm1.40}\) & \(\,82.17{\scriptscriptstyle \pm1.93}\) & \(\,46.48{\scriptscriptstyle \pm2.94}\) & \(\,41.91{\scriptscriptstyle \pm2.78}\) & \(\,41.41{\scriptscriptstyle \pm1.92}\) & \(\,50.22{\scriptscriptstyle \pm2.03}\) & \(\,66.84{\scriptscriptstyle \pm1.79}\) & \(\,89.43{\scriptscriptstyle \pm3.11}\) & \(\,46.06{\scriptscriptstyle \pm2.14}\) & \(\,62.76{\scriptscriptstyle \pm1.89}\) \\
      \midrule
      MMD~\cite{long2015learning} & \(\,75.18{\scriptscriptstyle \pm9.65}\) & \(\,64.45{\scriptscriptstyle \pm9.93}\) & \(\,36.44{\scriptscriptstyle \pm8.70}\) & \(\,81.63{\scriptscriptstyle \pm5.67}\) & \(\,64.73{\scriptscriptstyle \pm8.74}\) & \(\,41.06{\scriptscriptstyle \pm8.03}\) & \(\,50.50{\scriptscriptstyle \pm1.11}\) & \(\,50.69{\scriptscriptstyle \pm2.82}\) & \(\,67.04{\scriptscriptstyle \pm2.11}\) & \(\,80.60{\scriptscriptstyle \pm24.00}\) & \(\,74.33{\scriptscriptstyle \pm20.72}\) & \(\,75.16{\scriptscriptstyle \pm12.50}\) \\
      MMD-P & \(\,59.18{\scriptscriptstyle \pm5.84}\) & \(\,44.92{\scriptscriptstyle \pm3.03}\) & \(\,13.45{\scriptscriptstyle \pm1.10}\) & \(\,57.43{\scriptscriptstyle \pm3.55}\) & \(\,43.31{\scriptscriptstyle \pm4.61}\) & \(\,17.34{\scriptscriptstyle \pm7.92}\) & \(\,50.77{\scriptscriptstyle \pm1.14}\) & \(\,51.11{\scriptscriptstyle \pm2.07}\) & \(\,65.02{\scriptscriptstyle \pm4.89}\) & \(\,69.84{\scriptscriptstyle \pm17.90}\) & \(\,67.66{\scriptscriptstyle \pm17.99}\) & \(\,70.16{\scriptscriptstyle \pm15.44}\) \\
      DANN~\cite{ganin2016domain} & \(\,80.68{\scriptscriptstyle \pm2.18}\) & \(\,65.09{\scriptscriptstyle \pm3.83}\) & \(\,42.05{\scriptscriptstyle \pm4.29}\) & \(\,81.45{\scriptscriptstyle \pm1.54}\) & \(\,63.31{\scriptscriptstyle \pm2.61}\) & \(\,44.27{\scriptscriptstyle \pm1.04}\) & \(\,51.41{\scriptscriptstyle \pm3.16}\) & \(\,50.22{\scriptscriptstyle \pm2.03}\) & \(\,66.84{\scriptscriptstyle \pm1.79}\) & \(\,\textcolor{blue}{97.30{\scriptscriptstyle \pm2.78}}\) & \(\,\textcolor{blue}{94.52{\scriptscriptstyle \pm2.47}}\) & \(\,\textcolor{blue}{94.17{\scriptscriptstyle \pm2.57}}\) \\
      DANN-P & \(\,\textcolor{blue}{82.12{\scriptscriptstyle \pm2.61}}\) & \(\,66.61{\scriptscriptstyle \pm1.58}\) & \(\,\textcolor{blue}{44.53{\scriptscriptstyle \pm3.08}}\) & \(\,\textcolor{blue}{82.34{\scriptscriptstyle \pm1.15}}\) & \(\,65.60{\scriptscriptstyle \pm2.75}\) & \(\,\textcolor{blue}{46.66{\scriptscriptstyle \pm2.50}}\) & \(\,69.83{\scriptscriptstyle \pm11.29}\) & \(\,58.38{\scriptscriptstyle \pm11.43}\) & \(\,70.82{\scriptscriptstyle \pm6.40}\) & \(\,96.56{\scriptscriptstyle \pm2.99}\) & \(\,92.70{\scriptscriptstyle \pm4.42}\) & \(\,92.12{\scriptscriptstyle \pm4.84}\) \\
      CDAN~\cite{long2018conditional} & \(\,79.90{\scriptscriptstyle \pm1.84}\) & \(\,66.96{\scriptscriptstyle \pm1.52}\) & \(\,44.13{\scriptscriptstyle \pm1.69}\) & \(\,81.61{\scriptscriptstyle \pm2.47}\) & \(\,\textcolor{blue}{65.99{\scriptscriptstyle \pm1.92}}\) & \(\,44.88{\scriptscriptstyle \pm1.78}\) & \(\,57.14{\scriptscriptstyle \pm7.87}\) & \(\,50.22{\scriptscriptstyle \pm2.03}\) & \(\,66.84{\scriptscriptstyle \pm1.79}\) & \(\,86.36{\scriptscriptstyle \pm26.07}\) & \(\,85.09{\scriptscriptstyle \pm23.69}\) & \(\,88.34{\scriptscriptstyle \pm15.96}\) \\
      MCD~\cite{saito2018maximum} & \(\,80.71{\scriptscriptstyle \pm5.28}\) & \(\,\textcolor{blue}{67.05{\scriptscriptstyle \pm8.49}}\) & \(\,38.00{\scriptscriptstyle \pm8.16}\) & \(\,80.91{\scriptscriptstyle \pm1.77}\) & \(\,64.91{\scriptscriptstyle \pm4.88}\) & \(\,41.04{\scriptscriptstyle \pm4.24}\) & \(\,50.25{\scriptscriptstyle \pm0.15}\) & \(\,50.22{\scriptscriptstyle \pm2.03}\) & \(\,66.84{\scriptscriptstyle \pm1.79}\) & \(\,44.27{\scriptscriptstyle \pm21.14}\) & \(\,44.26{\scriptscriptstyle \pm2.40}\) & \(\,61.15{\scriptscriptstyle \pm1.73}\) \\
      SRDC~\cite{tang2020unsupervised} & \(\,64.84{\scriptscriptstyle \pm1.91}\) & \(\,49.19{\scriptscriptstyle \pm2.44}\) & \(\,17.72{\scriptscriptstyle \pm3.80}\) & \(\,69.57{\scriptscriptstyle \pm3.73}\) & \(\,45.57{\scriptscriptstyle \pm6.80}\) & \(\,24.42{\scriptscriptstyle \pm3.73}\) & \(\,50.36{\scriptscriptstyle \pm0.23}\) & \(\,50.22{\scriptscriptstyle \pm2.03}\) & \(\,66.84{\scriptscriptstyle \pm1.79}\) & \(\,17.76{\scriptscriptstyle \pm15.83}\) & \(\,46.75{\scriptscriptstyle \pm2.42}\) & \(\,49.90{\scriptscriptstyle \pm28.01}\) \\
      CORAL~\cite{sun2016deep} & \(\,76.27{\scriptscriptstyle \pm3.65}\) & \(\,55.80{\scriptscriptstyle \pm2.16}\) & \(\,25.90{\scriptscriptstyle \pm1.99}\) & \(\,80.72{\scriptscriptstyle \pm2.28}\) & \(\,44.25{\scriptscriptstyle \pm1.79}\) & \(\,33.73{\scriptscriptstyle \pm1.55}\) & \(\,57.45{\scriptscriptstyle \pm1.29}\) & \(\,50.22{\scriptscriptstyle \pm2.03}\) & \(\,66.84{\scriptscriptstyle \pm1.79}\) & \(\,57.60{\scriptscriptstyle \pm32.12}\) & \(\,58.17{\scriptscriptstyle \pm20.69}\) & \(\,68.74{\scriptscriptstyle \pm11.84}\) \\
      DRL{\scriptsize w/o label}~\cite{kim2026distributionally} & \(\,62.00{\scriptscriptstyle \pm14.33}\) & \(\,45.81{\scriptscriptstyle \pm10.60}\) & \(\,15.90{\scriptscriptstyle \pm6.35}\) & \(\,72.41{\scriptscriptstyle \pm6.58}\) & \(\,39.15{\scriptscriptstyle \pm16.10}\) & \(\,20.17{\scriptscriptstyle \pm13.11}\) & \(\,55.43{\scriptscriptstyle \pm3.38}\) & \(\,50.22{\scriptscriptstyle \pm2.03}\) & \(\,66.84{\scriptscriptstyle \pm1.79}\) & \(\,71.10{\scriptscriptstyle \pm31.83}\) & \(\,65.53{\scriptscriptstyle \pm22.86}\) & \(\,74.68{\scriptscriptstyle \pm13.31}\) \\
      DRL{\scriptsize w/i label} & \(\,72.40{\scriptscriptstyle \pm5.65}\) & \(\,51.97{\scriptscriptstyle \pm4.87}\) & \(\,20.21{\scriptscriptstyle \pm6.85}\) & \(\,77.69{\scriptscriptstyle \pm1.26}\) & \(\,61.08{\scriptscriptstyle \pm5.64}\) & \(\,38.10{\scriptscriptstyle \pm8.42}\) & \(\,51.67{\scriptscriptstyle \pm3.42}\) & \(\,50.22{\scriptscriptstyle \pm2.03}\) & \(\,66.84{\scriptscriptstyle \pm1.79}\) & \(\,99.97{\scriptscriptstyle \pm0.06}\) & \(\,99.13{\scriptscriptstyle \pm0.58}\) & \(\,99.47{\scriptscriptstyle \pm0.45}\) \\
      SimProF~\cite{wang2025simprof}
      & \(\,79.13{\scriptscriptstyle \pm2.98}\)
      & \(\,50.17{\scriptscriptstyle \pm11.58}\)
      & \(\,41.53{\scriptscriptstyle \pm5.34}\)
      & \(\,80.99{\scriptscriptstyle \pm2.43}\)
      & \(\,52.30{\scriptscriptstyle \pm7.12}\)
      & \(\,44.96{\scriptscriptstyle \pm5.52}\)
      & \(\,\textcolor{blue}{78.41{\scriptscriptstyle \pm13.82}}\)
      & \(\,\textcolor{blue}{70.31{\scriptscriptstyle \pm12.79}}\)
      & \(\,\textcolor{red}{77.58{\scriptscriptstyle \pm7.49}}\)
      & \(\,92.76{\scriptscriptstyle \pm2.77}\)
      & \(\,87.67{\scriptscriptstyle \pm12.62}\)
      & \(\,88.63{\scriptscriptstyle \pm9.71}\) \\
      \midrule
      DETA~\cite{shou2025graph} & \(\,65.73{\scriptscriptstyle \pm3.90}\) & \(\,46.97{\scriptscriptstyle \pm0.60}\) & \(\,13.26{\scriptscriptstyle \pm1.10}\) & \(\,63.85{\scriptscriptstyle \pm9.71}\) & \(\,40.62{\scriptscriptstyle \pm15.88}\) & \(\,16.46{\scriptscriptstyle \pm9.74}\) & \(\,44.46{\scriptscriptstyle \pm5.66}\) & \(\,50.26{\scriptscriptstyle \pm2.09}\) & \(\,66.86{\scriptscriptstyle \pm1.82}\) & \(\,38.13{\scriptscriptstyle \pm7.66}\) & \(\,47.52{\scriptscriptstyle \pm1.57}\) & \(\,64.41{\scriptscriptstyle \pm1.38}\) \\
      RAM-MIL~\cite{cui2023retrieval} & \(\,76.78{\scriptscriptstyle \pm1.31}\) & \(\,56.60{\scriptscriptstyle \pm1.85}\) & \(\,28.58{\scriptscriptstyle \pm2.08}\) & \(\,81.55{\scriptscriptstyle \pm2.16}\) & \(\,58.50{\scriptscriptstyle \pm0.97}\) & \(\,31.21{\scriptscriptstyle \pm0.45}\) & \(\,57.29{\scriptscriptstyle \pm1.42}\) & \(\,50.22{\scriptscriptstyle \pm2.03}\) & \(\,66.84{\scriptscriptstyle \pm1.79}\) & \(\,89.34{\scriptscriptstyle \pm2.30}\) & \(\,44.93{\scriptscriptstyle \pm1.83}\) & \(\,61.86{\scriptscriptstyle \pm1.63}\) \\
      \midrule
      ours & \(\,\textcolor{red}{83.89{\scriptscriptstyle \pm1.42}}\) & \(\,\textcolor{red}{67.50{\scriptscriptstyle \pm0.56}}\) & \(\,\textcolor{red}{46.83{\scriptscriptstyle \pm2.82}}\) & \(\,\textcolor{red}{83.45{\scriptscriptstyle \pm0.68}}\) & \(\,\textcolor{red}{67.09{\scriptscriptstyle \pm2.31}}\) & \(\,\textcolor{red}{47.63{\scriptscriptstyle \pm1.76}}\) & \(\,\textcolor{red}{87.39{\scriptscriptstyle \pm6.01}}\) & \(\,\textcolor{red}{77.71{\scriptscriptstyle \pm6.74}}\) & \(\,\textcolor{blue}{75.97{\scriptscriptstyle \pm9.84}}\) & \(\,\textcolor{red}{97.83{\scriptscriptstyle \pm1.10}}\) & \(\,\textcolor{red}{94.71{\scriptscriptstyle \pm2.27}}\) & \(\,\textcolor{red}{94.39{\scriptscriptstyle \pm2.41}}\) \\
      \bottomrule
    \end{tabular}
  }
\end{table}

To assess DARE's plug-and-play applicability, we evaluate it and the compared
adaptation methods using CLAM~\citep{lu2021data} as the shared backbone.
DARE achieves the highest AUC across all four transfer directions
(Table~\ref{tab:plug_play_clam}), supporting its applicability beyond
DTFD-MIL to attention-based MIL backbones that provide patch-importance scores.

\begin{table}[t]
\centering
\caption{AUC (\%, mean $\pm$ standard deviation) using CLAM as the shared MIL backbone.}
\label{tab:plug_play_clam}
\setlength{\tabcolsep}{3pt}
\begin{tabular}{lcccc}
\toprule
Method
& \shortstack{TCGA $\rightarrow$ IPD\\(GBMLGG)}
& \shortstack{IPD $\rightarrow$ TCGA\\(GBMLGG)}
& \shortstack{CPTAC $\rightarrow$ TCGA\\(NSCLC)}
& \shortstack{TCGA $\rightarrow$ CPTAC\\(NSCLC)} \\
\midrule
DANN
& $80.49 \pm {\scriptscriptstyle 4.92}$
& $80.77 \pm {\scriptscriptstyle 2.94}$
& $50.00 \pm {\scriptscriptstyle 0.00}$
& $95.70 \pm {\scriptscriptstyle 7.58}$ \\
DANN-P
& $72.64 \pm {\scriptscriptstyle 4.26}$
& $75.64 \pm {\scriptscriptstyle 1.85}$
& $62.24 \pm {\scriptscriptstyle 5.01}$
& $35.34 \pm {\scriptscriptstyle 7.02}$ \\
MCD
& $80.33 \pm {\scriptscriptstyle 3.05}$
& $81.24 \pm {\scriptscriptstyle 1.62}$
& $50.43 \pm {\scriptscriptstyle 0.44}$
& $85.40 \pm {\scriptscriptstyle 29.44}$ \\
CORAL
& $77.57 \pm {\scriptscriptstyle 2.43}$
& $82.28 \pm {\scriptscriptstyle 1.67}$
& $62.42 \pm {\scriptscriptstyle 10.77}$
& $60.36 \pm {\scriptscriptstyle 24.37}$ \\
RAM-MIL
& $80.27 \pm {\scriptscriptstyle 1.20}$
& $82.42 \pm {\scriptscriptstyle 1.97}$
& $37.42 \pm {\scriptscriptstyle 2.29}$
& $88.67 \pm {\scriptscriptstyle 2.94}$ \\
Ours
& \textcolor{red}{$84.83 \pm {\scriptscriptstyle 1.65}$}
& \textcolor{red}{$83.01 \pm {\scriptscriptstyle 2.52}$}
& \textcolor{red}{$84.11 \pm {\scriptscriptstyle 9.92}$}
& \textcolor{red}{$99.57 \pm {\scriptscriptstyle 0.33}$} \\
\bottomrule
\end{tabular}
\end{table}

\subsection{Analysis of Feature Extraction Backbones}
In the main experiments, all methods utilize patch features extracted by the CONCH encoder.
To evaluate whether the proposed DARE generalizes well to other patch feature extractors, we further conduct experiments using the foundation model UNI2‑h~\cite{chen2024uni} as the encoder for feature extraction.
The results, summarized in Table~\ref{tab:encoder_robustness}, show that DARE consistently outperforms the compared methods under this feature extractor, demonstrating its robustness to the choice of the underlying encoder.

\begin{table}[t]
  \centering
  \begin{minipage}[c]{0.6\textwidth}
    \centering
    \caption{Experimental results based on the UNI2-h encoder on two cross-domain settings.}
    \label{tab:encoder_robustness}
    \setlength{\tabcolsep}{1.25pt}
    \resizebox{\linewidth}{!}{
      \begin{tabular}{*{7}{c}}
        \toprule
        \multirow{2}{*}{methods} & \multicolumn{3}{c}{TCGA \(\rightarrow\) IPD (GBMLGG)} & \multicolumn{3}{c}{IPD \(\rightarrow\) TCGA (GBMLGG)} \\
        \cmidrule(lr){2-4}\cmidrule(lr){5-7}
        & AUC & ACC & F1 & AUC & ACC & F1 \\
        \midrule
        DTFD & \(\,84.92{\scriptscriptstyle \pm2.59}\) & \(\,73.24{\scriptscriptstyle \pm1.49}\) & \(\,49.61{\scriptscriptstyle \pm2.53}\) & \(\,85.22{\scriptscriptstyle \pm1.40}\) & \(\,71.84{\scriptscriptstyle \pm0.96}\) & \(\,\textcolor{blue}{48.35{\scriptscriptstyle \pm0.99}}\) \\
        DANN & \(\,84.17{\scriptscriptstyle \pm3.36}\) & \(\,69.04{\scriptscriptstyle \pm2.10}\) & \(\,46.36{\scriptscriptstyle \pm2.42}\) & \(\,82.52{\scriptscriptstyle \pm2.51}\) & \(\,67.80{\scriptscriptstyle \pm5.93}\) & \(\,46.89{\scriptscriptstyle \pm2.54}\) \\
        DANN-P & \(\,81.24{\scriptscriptstyle \pm5.44}\) & \(\,72.79{\scriptscriptstyle \pm5.90}\) & \(\,48.46{\scriptscriptstyle \pm4.86}\) & \(\,84.36{\scriptscriptstyle \pm2.00}\) & \(\,71.11{\scriptscriptstyle \pm1.95}\) & \(\,48.17{\scriptscriptstyle \pm2.57}\) \\
        CORAL & \(\,83.65{\scriptscriptstyle \pm2.70}\) & \(\,\textcolor{blue}{74.22{\scriptscriptstyle \pm2.64}}\) & \(\,48.48{\scriptscriptstyle \pm1.59}\) & \(\,\textcolor{blue}{85.23{\scriptscriptstyle \pm1.35}}\) & \(\,71.47{\scriptscriptstyle \pm1.11}\) & \(\,48.23{\scriptscriptstyle \pm1.33}\) \\
        MCD & \(\,\textcolor{blue}{86.37{\scriptscriptstyle \pm4.47}}\) & \(\,69.04{\scriptscriptstyle \pm2.72}\) & \(\,\textcolor{blue}{49.68{\scriptscriptstyle \pm6.11}}\) & \(\,83.51{\scriptscriptstyle \pm2.70}\) & \(\,70.11{\scriptscriptstyle \pm3.80}\) & \(\,47.56{\scriptscriptstyle \pm2.53}\) \\
        RAM-MIL & \(\,85.29{\scriptscriptstyle \pm2.52}\) & \(\,72.97{\scriptscriptstyle \pm1.40}\) & \(\,48.64{\scriptscriptstyle \pm1.79}\) & \(\,84.28{\scriptscriptstyle \pm1.57}\) & \(\,\textcolor{blue}{71.87{\scriptscriptstyle \pm1.03}}\) & \(\,46.28{\scriptscriptstyle \pm0.95}\) \\
        ours & \(\,\textcolor{red}{86.80{\scriptscriptstyle \pm2.94}}\) & \(\,\textcolor{red}{75.64{\scriptscriptstyle \pm2.41}}\) & \(\,\textcolor{red}{51.21{\scriptscriptstyle \pm2.46}}\) & \(\,\textcolor{red}{86.17{\scriptscriptstyle \pm1.89}}\) & \(\,\textcolor{red}{72.23{\scriptscriptstyle \pm2.85}}\) & \(\,\textcolor{red}{53.05{\scriptscriptstyle \pm3.99}}\) \\
        \bottomrule
      \end{tabular}
    }
  \end{minipage}\hfill
  \begin{minipage}[c]{0.38\textwidth}
    \centering
    \caption{Ablation study results on two datasets.}
    \label{tab:ablation_results}
    \setlength{\tabcolsep}{1.5pt}
    \resizebox{\linewidth}{!}{
      \begin{tabular}{lcc}
        \toprule
        method & TCGA \(\rightarrow\) IPD & IPD \(\rightarrow\) TCGA \\
        \midrule
        w/o domain-aware & 82.90 & 81.78 \\
        w/o \(\mathcal{L}_{ins}\)     & 72.49 & 80.09 \\
        w/o \(\mathcal{L}_{slide}\)   & 75.92 & 76.52 \\
        w/o \(\mathcal{L}_{pseudo}\)  & 83.58 & 79.74 \\
        \midrule
        \textcolor{red}{ours} & \textcolor{red}{83.89} & \textcolor{red}{83.45} \\
        \bottomrule
      \end{tabular}
    }
  \end{minipage}
\end{table}

\subsection{Ablation Studies}
We perform ablation studies on the TCGA-GBMLGG and IPD-Brain datasets. 
As shown in Table~\ref{tab:ablation_results}, we denote each ablation setting as follows: “w/o domain-aware” removes domain probability guidance masking; “w/o $\mathcal{L}_{ins}$” removes patch-level domain adversarial learning; “w/o $\mathcal{L}_{slide}$” removes slide-level domain adversarial learning; and “w/o $\mathcal{L}_{pseudo}$” removes pseudo-label supervision on the target domain. The full model achieves the best performance.
Both patch-level and slide-level domain adversarial modules provide significant gains, highlighting the importance of multi-granularity embedding alignment. The domain-aware masking also yields consistent improvement. Notably, on the IPD $\rightarrow$ TCGA task, pseudo-labeling leads to substantial performance gains, reinforcing its effectiveness in providing supervision signals for the target domain. Overall, each module contributes positively to the overall performance, validating the effectiveness of every component in our framework.

\subsection{Robustness to Limited Target Domain Data}
In real-world clinical scenarios, some hospitals and clinical centers may face difficulties in obtaining large-scale datasets for model adaptation. To evaluate the data efficiency of our method under such resource-limited conditions, we conduct experiments with limited target data by varying the proportion of target domain data used for adaptation.
As shown in Table~\ref{tab:few_shot_results}, the proposed method delivers remarkable gains even with extremely limited target domain data. 
With only 5\% of the target domain data, both ACC and F1 improve substantially over the performance without any adaptation, confirming the strong data efficiency of our approach in adapting to new domains with minimal unlabeled target samples.

\begin{table}[htbp]
  \centering
  \caption{Robustness to Limited Target Domain Data. Ratio: proportion of target data used. Eval: without adaptation.}
  \label{tab:few_shot_results}
  \setlength{\tabcolsep}{1.25pt}
  \resizebox{\textwidth}{!}{
    \begin{tabular}{c*{12}{c}}
      \toprule
      \multirow{2}{*}{ratio} & \multicolumn{3}{c}{TCGA \(\rightarrow\) IPD (GBMLGG)} & \multicolumn{3}{c}{IPD \(\rightarrow\) TCGA (GBMLGG)} & \multicolumn{3}{c}{CPTAC \(\rightarrow\) TCGA (NSCLC)} & \multicolumn{3}{c}{TCGA \(\rightarrow\) CPTAC (NSCLC)} \\
      \cmidrule(lr){2-4}\cmidrule(lr){5-7}\cmidrule(lr){8-10}\cmidrule(lr){11-13}
      & AUC & ACC & F1 & AUC & ACC & F1 & AUC & ACC & F1 & AUC & ACC & F1 \\
      \midrule
      eval & \(73.74{\scriptscriptstyle\pm0.58}\) & \(53.66{\scriptscriptstyle\pm0.64}\) & \(25.18{\scriptscriptstyle\pm3.18}\) & \(80.48{\scriptscriptstyle\pm2.41}\) & \(43.00{\scriptscriptstyle\pm2.04}\) & \(29.07{\scriptscriptstyle\pm1.61}\) & \(57.46{\scriptscriptstyle\pm1.30}\) & \(50.22{\scriptscriptstyle\pm2.03}\) & \(66.85{\scriptscriptstyle\pm1.80}\) & \(87.43{\scriptscriptstyle\pm2.93}\) & \(54.24{\scriptscriptstyle\pm1.44}\) & \(65.69{\scriptscriptstyle\pm1.22}\) \\
      5\%    & \(80.38{\scriptscriptstyle\pm4.08}\) & \(59.29{\scriptscriptstyle\pm8.84}\) & \(42.68{\scriptscriptstyle\pm7.73}\) & \(82.11{\scriptscriptstyle\pm1.81}\) & \(65.99{\scriptscriptstyle\pm1.91}\) & \(45.90{\scriptscriptstyle\pm2.25}\) & \(74.99{\scriptscriptstyle\pm13.45}\) & \(68.00{\scriptscriptstyle\pm11.92}\) & \(70.22{\scriptscriptstyle\pm10.23}\) & \(92.99{\scriptscriptstyle\pm3.74}\) & \(79.83{\scriptscriptstyle\pm16.18}\) & \(82.94{\scriptscriptstyle\pm11.12}\) \\
      10\%   & \(83.53{\scriptscriptstyle\pm3.41}\) & \(65.10{\scriptscriptstyle\pm4.34}\) & \(44.64{\scriptscriptstyle\pm2.73}\) & \(80.74{\scriptscriptstyle\pm3.34}\) & \(64.39{\scriptscriptstyle\pm3.96}\) & \(44.73{\scriptscriptstyle\pm3.80}\) & \(81.32{\scriptscriptstyle\pm9.36}\) & \(74.90{\scriptscriptstyle\pm9.15}\) & \(77.25{\scriptscriptstyle\pm8.53}\) & \(93.72{\scriptscriptstyle\pm6.32}\) & \(85.15{\scriptscriptstyle\pm13.33}\) & \(86.67{\scriptscriptstyle\pm10.64}\) \\
      15\%   & \(82.91{\scriptscriptstyle\pm2.14}\) & \(65.09{\scriptscriptstyle\pm3.79}\) & \(47.22{\scriptscriptstyle\pm3.96}\) & \(81.98{\scriptscriptstyle\pm2.59}\) & \(65.12{\scriptscriptstyle\pm2.43}\) & \(45.64{\scriptscriptstyle\pm2.42}\) & \(87.20{\scriptscriptstyle\pm7.19}\) & \(78.26{\scriptscriptstyle\pm8.65}\) & \(76.72{\scriptscriptstyle\pm9.76}\) & \(91.27{\scriptscriptstyle\pm9.11}\) & \(87.54{\scriptscriptstyle\pm13.70}\) & \(88.83{\scriptscriptstyle\pm10.15}\) \\
      \bottomrule
    \end{tabular}
  }
\end{table}

\subsection{Visualization of the Feature Alignment}
\begin{figure}[t]
  \centering
  \includegraphics[width=\linewidth]{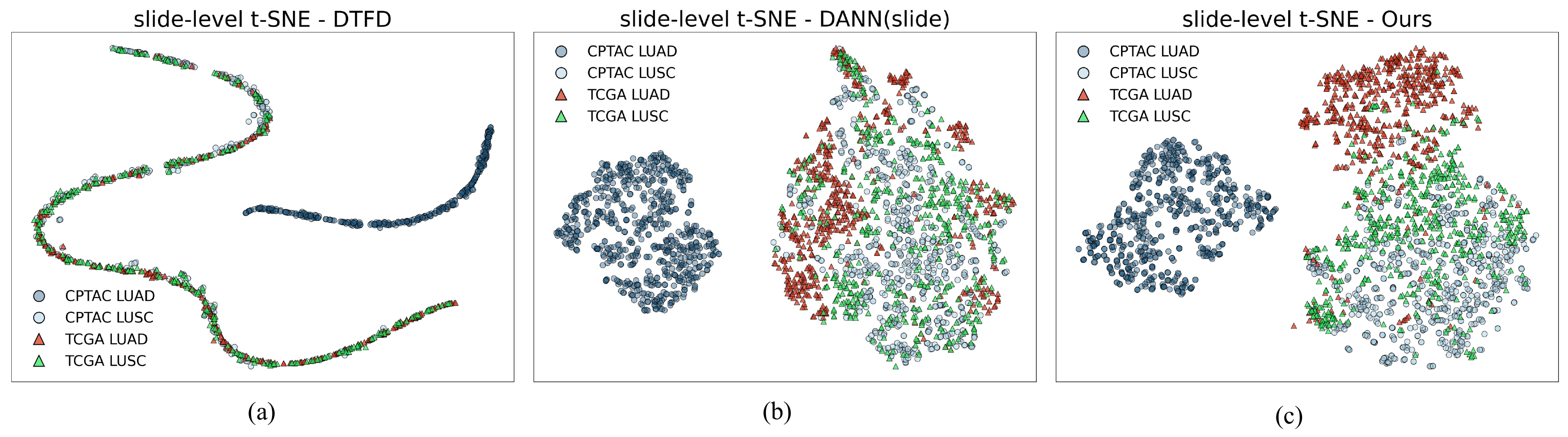}
  \caption{T-SNE visualization of source and target embedding distributions for the CPTAC \(\rightarrow\) TCGA adaptation, comparing DTFD, DANN, and our method.}
  \label{fig:tsne_visualization}
\end{figure}

To demonstrate the effectiveness of the dual-level feature alignment, we visualize the feature distributions of the source and target domains of different methods. Fig.~\ref{fig:tsne_visualization} presents the slide-level embeddings of three methods: (a) the original DTFD backbone, (b) DANN (slide-level) with domain adversarial training, and (c) our proposed method.
The embedding space of the original DTFD exhibits two distinct manifolds with compact intra-cluster structures.
The two target subtypes are nearly indistinguishable in the embedding space.
DANN reduces the domain gap across domains. However, the features of the two subtypes in the target domain still overlap in several regions, making them difficult to distinguish.
In contrast, our method reduces domain discrepancy while producing well-separated clusters for target domain subtypes, leading to clearer decision boundaries and improved discrimination over the baselines.

\section{Conclusions}
We propose DARE, a plug-and-play framework for unsupervised domain adaptation in computational pathology. The model addresses the hierarchical relationship between patches and slides through a dual-level domain adversarial mechanism for multi-granularity embedding alignment. We further introduce a domain-aware attention masking strategy that regularizes MIL attention and facilitates hard sample mining across domains. In addition, we design a teacher-student-based adaptive pseudo-labeling mechanism to provide supervision for target-domain data and enable class-wise alignment. We evaluate DARE on four datasets and demonstrate its effectiveness. Extensive comparisons with state-of-the-art methods show that DARE achieves superior performance in unsupervised domain adaptation for pathology images.

\begin{ack}
This work was supported by the National Key Research and Development Program of China (Grant No. 2026YFE0199900) and the National Natural Science Foundation of China (Grant No. 62471149). Additional support was provided by the CFFF Intelligent Computing Platform at Fudan University.
\end{ack}

\bibliographystyle{plainnat}
\bibliography{NIPS}

\newpage
\appendix

\section{Dataset Description}
We conduct experiments on four datasets: TCGA-NSCLC, CPTAC-NSCLC, TCGA-GBMLGG, and IPD-Brain.

\textbf{TCGA-NSCLC dataset} contains 1,040 WSIs for binary classification of Lung Adenocarcinoma (LUAD) versus Lung Squamous Cell Carcinoma (LUSC).

\textbf{CPTAC-NSCLC dataset} contains 1,342 WSIs and is also used for LUAD versus LUSC classification. Compared with TCGA-NSCLC, it shows a clear shift in staining appearance and a noticeable domain gap in visual distribution.

\textbf{TCGA-GBMLGG dataset} contains 1,526 WSIs, integrating both lower-grade glioma and glioblastoma cases. Based on IDH1 mutation status and histopathological subtypes, samples are categorized into five molecular subtypes: IDH1-mutant astrocytoma, IDH1-mutant oligodendroglioma, IDH1-mutant glioblastoma, IDH1-wildtype astrocytoma, and IDH1-wildtype glioblastoma. In our experiments, it is used for a five-class classification task. Due to large morphological variability across subtypes and heterogeneous data sources, this dataset exhibits strong intrinsic distributional heterogeneity.

\textbf{IPD-Brain dataset} contains 448 WSIs and is collected from an Indian glioma cohort. It is used for the same five-class classification task. Compared with TCGA-GBMLGG, it shows a clear distribution shift in patient population. This cross-region variation further increases the difficulty of feature transfer and decision boundary alignment.

\section{Implementation Details}
\label{sec:Implementation Details}

For our DARE, we follow the standard unsupervised domain adaptation protocol: we first pre-train a source model under the DTFD framework on the labeled source domain, and then load this pre-trained model as initialization for domain adaptation on the combined labeled source and unlabeled target data. 

Specifically, we use the Adam optimizer with a learning rate of \(1\times10^{-4}\) and weight decay of \(1\times10^{-4}\).
For the dual-level domain adversarial module, the loss weights default to \(\lambda_{slide} = 0.5\) and \(\lambda_{ins} = 0.7\).
For the domain-aware attention masking module, the masking ratio \(r\) defaults to \(0.1\).
For the teacher-student pseudo-labeling module, the exponential moving average smoothing coefficient for the adaptive threshold defaults to \(\alpha = 0.8\), with a maximum allowed change \(\Delta = 0.05\) and a clipping range \([\tau_{\min},\tau_{\max}] = [0.7, 0.95]\). The teacher model parameters \(\theta_t\) are updated via \(\theta_t \leftarrow \alpha_{t} \theta_t + (1 - \alpha_{t}) \theta_s\), where \(\theta_s\) denotes the student parameters and \(\alpha_{t}=0.99\) by default. The pseudo-label loss weight defaults to \(\lambda_{pseudo} = 0.5\). To avoid division by zero when computing the class-specific normalized distillation weights \(w_c\), we set a minimum of 4 accepted samples per class (\(|A_c| \ge 4\)). Pseudo-label learning is activated at the 40th epoch as a warm-up phase to mitigate confirmation bias from unreliable early-stage predictions.
All experiments are conducted on an NVIDIA A100 Tensor Core GPU.

\section{Hyperparameter Analysis}
\begin{figure}[htbp]
  \centering
  \includegraphics[width=0.8\linewidth]{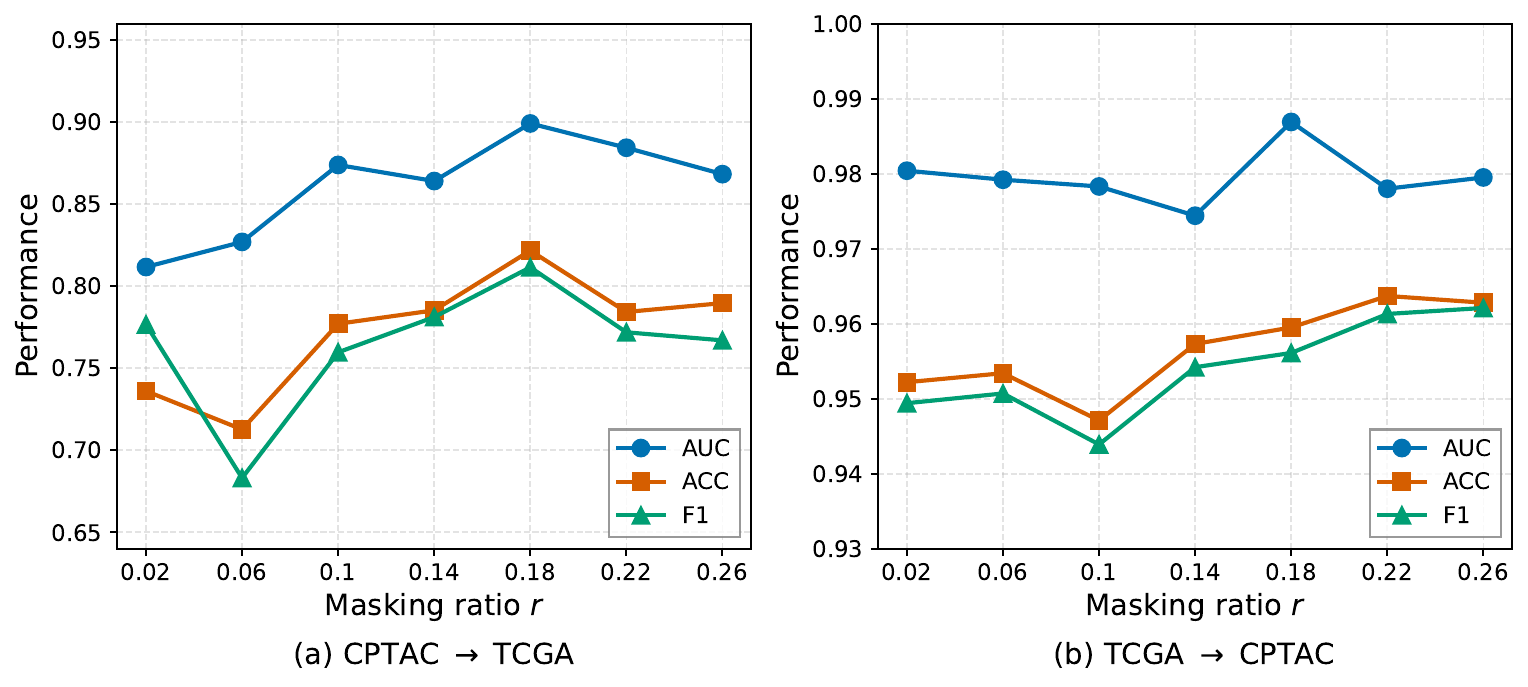}
  \caption{Sensitivity analysis of the masking ratio \(r\) on two domain adaptation tasks.
  (a) CPTAC \(\rightarrow\) TCGA.
  (b) TCGA \(\rightarrow\) CPTAC.
  Performance is measured by ACC, F1-score, and AUC.}
  \label{fig:masking_ratio}
\end{figure}

We conduct hyperparameter experiments to explore the impact of different values for the masking ratio \(r\) on the CPTAC-NSCLC and TCGA-NSCLC datasets.
As shown in Fig.~\ref{fig:masking_ratio}, on the harder CPTAC \(\rightarrow\) TCGA task, all three metrics improve substantially as \(r\) increases from 0.02 to 0.18, confirming that suppressing domain-specific easy patches effectively regularizes attention and facilitates hard sample mining. The peak AUC is reached at \(r=0.18\). When \(r\) continues to increase beyond 0.18, performance gradually declines (AUC drops from 0.8992 to 0.8844 at \(r=0.22\) and further to 0.8683 at \(r=0.26\)), suggesting that over-aggressive masking begins to discard useful discriminative patches. In contrast, on the easier TCGA \(\rightarrow\) CPTAC task, the model is robust to \(r\) and maintains high performance across the entire tested range from 0.02 to 0.26, indicating that the domain‑aware masking does not hurt transfer when the domain gap is small. We use \(r=0.1\) as the default setting.

We further examine the sensitivity of the pseudo-label learning module to
$\tau_{\min}$, $\tau_{\max}$, $\alpha$, and $\Delta$ on the two GBMLGG transfer
tasks. We vary one parameter at a time while keeping the remaining settings
fixed. Table~\ref{tab:pseudo_hyperparameters} reports the mean AUC over
multiple random seeds, following the protocol used in the main experiments.

For $\tau_{\min}$, both transfer directions maintain competitive performance
over a range of values. The default value of $0.70$ achieves the highest AUC
on IPD-Brain $\rightarrow$ TCGA-GBMLGG, whereas $0.80$ performs best in the
reverse direction. The results for $\tau_{\max}$ are stable between $0.93$
and $0.99$ on TCGA-GBMLGG $\rightarrow$ IPD-Brain; the default value of $0.95$
performs best in the reverse direction. For $\alpha$, different values favor
the two transfer directions, and the default value of $0.80$ provides a
balanced choice. Setting $\Delta$ to $0.05$ yields the highest AUC in both
directions, supporting the use of adaptive threshold updates. Overall, the
default configuration performs competitively across both transfer directions
without tuning parameters separately for each task.

\begin{table}[t]
\centering
\caption{Sensitivity of the pseudo-label learning module to
$\tau_{\min}$, $\tau_{\max}$, $\alpha$, and $\Delta$. Results are mean AUC
(\%). One parameter is varied at a time; default values are indicated
in parentheses. The best result within each parameter group and transfer
direction is shown in red.}
\label{tab:pseudo_hyperparameters}
\small
\setlength{\tabcolsep}{6pt}
\begin{tabular}{llcc}
\toprule
Parameter & Value
& \shortstack{TCGA-GBMLGG $\rightarrow$\\IPD-Brain}
& \shortstack{IPD-Brain $\rightarrow$\\TCGA-GBMLGG} \\
\midrule
$\tau_{\min}$ & 0.60 & 82.34 & 82.11 \\
& 0.64 & 84.72 & 82.81 \\
& 0.67 & 83.47 & 82.69 \\
& 0.70 (default) & 83.89 & \textcolor{red}{83.45} \\
& 0.74 & 84.21 & 83.16 \\
& 0.77 & 84.26 & 82.45 \\
& 0.80 & \textcolor{red}{85.45} & 81.99 \\
\midrule
$\tau_{\max}$ & 0.80 & 81.35 & 81.39 \\
& 0.90 & \textcolor{red}{84.12} & 80.36 \\
& 0.93 & 83.89 & 83.12 \\
& 0.95 (default) & 83.89 & \textcolor{red}{83.45} \\
& 0.99 & 83.89 & 83.02 \\
\midrule
$\alpha$ & 0.75 & 84.33 & 82.72 \\
& 0.80 (default) & 83.89 & \textcolor{red}{83.45} \\
& 0.85 & 83.06 & 82.27 \\
& 0.90 & \textcolor{red}{85.28} & 82.36 \\
\midrule
$\Delta$ & 0.00 & 82.11 & 81.41 \\
& 0.05 (default) & \textcolor{red}{83.89} & \textcolor{red}{83.45} \\
& 0.10 & 81.97 & 83.06 \\
\bottomrule
\end{tabular}
\end{table}

\section{Pseudo-label Learning Ablation}
We assess three design choices in the pseudo-label learning module on the
five-class TCGA-GBMLGG and IPD-Brain datasets: adaptive threshold selection,
class-balanced weighting, and the KL-divergence-based pseudo-label loss.
For the loss comparison, we replace the KL-divergence objective with
cross-entropy against a one-hot label predicted by the teacher. All other
settings are kept the same as in the full DARE model.

As shown in Table~\ref{tab:pseudo_label_ablation}, removing adaptive
threshold selection decreases AUC in both transfer directions. Removing
class-balanced weighting causes a larger decrease, indicating its value
for learning from target-domain pseudo-labels. Replacing the KL-divergence
loss with cross-entropy against a one-hot teacher label also reduces AUC
in both directions. The full model achieves the highest AUC across these
comparisons.

\begin{table}[t]
\centering
\caption{Ablation study of pseudo-label learning on the two GBMLGG
transfer tasks. Results are AUC (\%, mean $\pm$ standard deviation).
The best result in each column is shown in red.}
\label{tab:pseudo_label_ablation}
\setlength{\tabcolsep}{5pt}
\resizebox{0.9\linewidth}{!}{%
\begin{tabular}{lcc}
\toprule
Method
& TCGA-GBMLGG $\rightarrow$ IPD-Brain
& IPD-Brain $\rightarrow$ TCGA-GBMLGG \\
\midrule
DARE (full)
& \textcolor{red}{$83.89{\scriptscriptstyle \pm 1.42}$}
& \textcolor{red}{$83.45{\scriptscriptstyle \pm 0.68}$} \\
w/o adaptive threshold selection
& $82.11{\scriptscriptstyle \pm 4.25}$
& $81.41{\scriptscriptstyle \pm 3.76}$ \\
w/o class-balanced weighting
& $79.07{\scriptscriptstyle \pm 5.33}$
& $78.84{\scriptscriptstyle \pm 1.93}$ \\
Cross-entropy with one-hot teacher labels
& $82.71{\scriptscriptstyle \pm 3.97}$
& $82.58{\scriptscriptstyle \pm 2.02}$ \\
\bottomrule
\end{tabular}%
}
\end{table}

\section{Visualization of Attention and Domain-Aware Masking}

\subsection{Visualization of Attention Redistribution}

\begin{figure}[htbp]
  \centering
  \includegraphics[width=\linewidth]{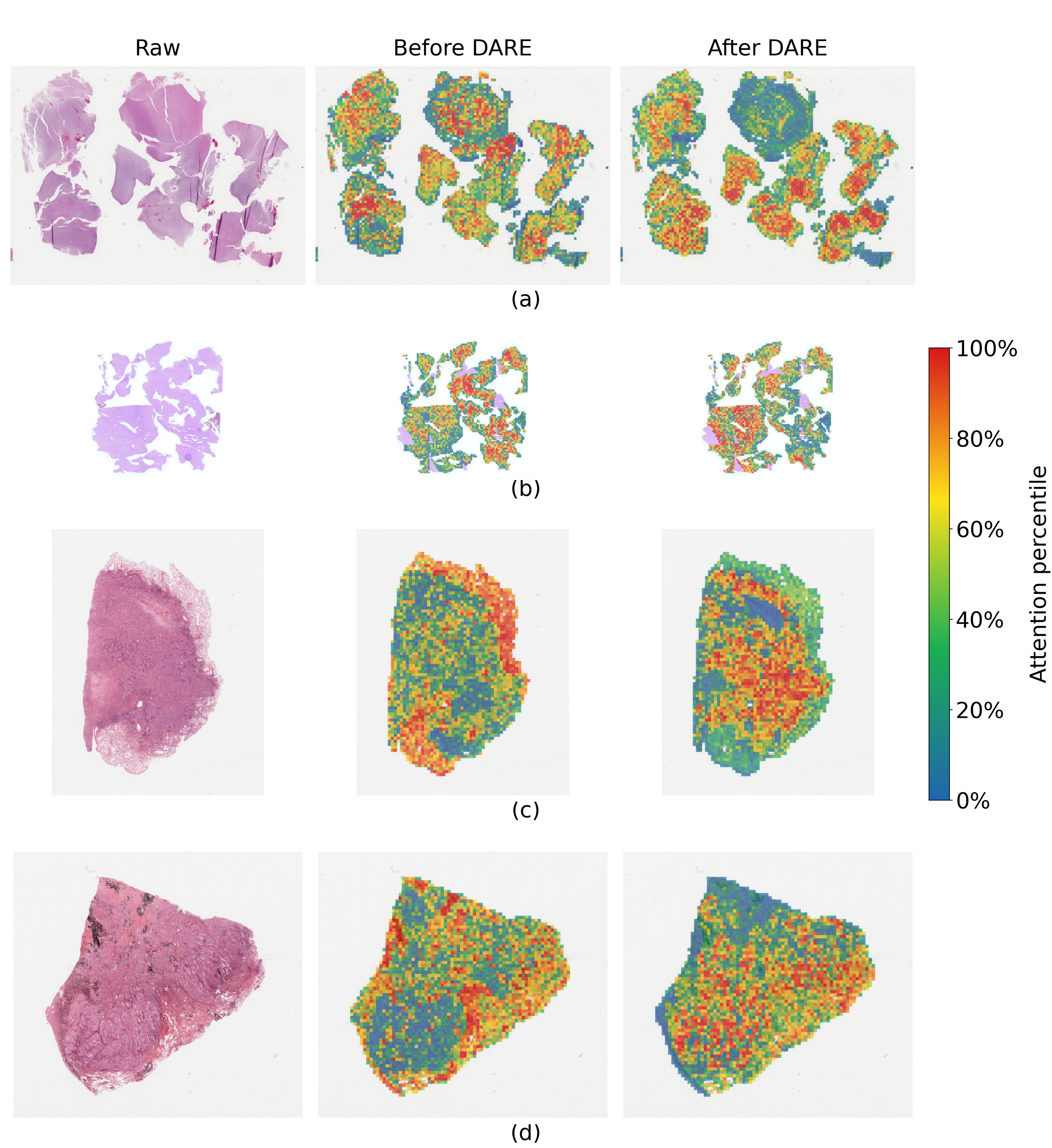}
  \caption{Attention maps before and after DARE for (a,b) glioma subtyping and (c,d) lung cancer subtyping.}
  \label{fig:attention_analysis}
\end{figure}

To examine how DARE affects the spatial distribution of attention, Fig.~\ref{fig:attention_analysis} presents representative target-domain slides for glioma and lung cancer subtyping. Each example includes the original H\&E image and attention maps before and after adaptation. Colors indicate attention percentiles, reflecting the relative ranking of patch attention.

For glioma subtyping (Fig.~\ref{fig:attention_analysis}a,b), DARE redistributes attention toward more cellular regions. In Fig.~\ref{fig:attention_analysis}a, the upper-central tissue fragment receives lower relative attention after adaptation, whereas regions including the lower-left fragment receive greater attention. Local H\&E inspection indicates that the former region is relatively sparsely cellular and contains neuron-like structures, while the latter regions exhibit more densely packed small, round nuclei. Similarly, in Fig.~\ref{fig:attention_analysis}b, relative attention increases in the lower-left region containing dense small, round nuclei and decreases in portions of the right-sided tissue with more prominent neuron-like cells and lower cellularity. These patterns suggest reduced emphasis on some background brain parenchymal features and increased attention to potentially tumor-associated cellular regions.

For lung cancer subtyping (Fig.~\ref{fig:attention_analysis}c,d), attention shifts toward regions exhibiting adenocarcinoma-associated morphology. In Fig.~\ref{fig:attention_analysis}c, high-ranking attention shifts from portions of the peripheral tissue with preserved alveolar architecture toward internal regions containing irregular gland-like structures. In Fig.~\ref{fig:attention_analysis}d, the upper fibrous region containing dark granular deposits receives lower relative attention, while the lower-left region with irregular gland-like structures receives greater attention.

Overall, these examples suggest that DARE redistributes relative attention toward potentially task-relevant morphology while reducing emphasis on some background tissue components.

\subsection{Visualization of Domain-Aware Masking}

\begin{figure}[htbp]
  \centering
  \includegraphics[width=\linewidth]{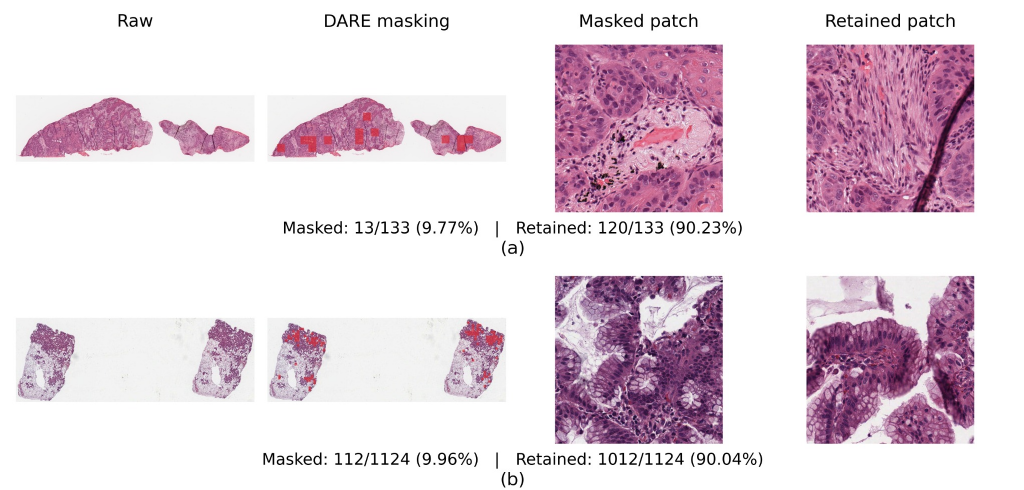}
  \caption{Domain-aware masking in lung cancer subtyping. Columns show the original H\&E images, masking maps, and attention-matched masked and retained patches. Red overlays indicate masked regions.}
  \label{fig:masking_analysis}
\end{figure}

To examine the behavior of domain-aware attention masking, Fig.~\ref{fig:masking_analysis} presents representative target-domain slides together with masked and retained patches of comparable attention. This comparison highlights how source-domain affinity contributes to patch selection beyond attention alone.

In Fig.~\ref{fig:masking_analysis}a, both the masked and retained patches contain epithelial-like cell groups interspersed with stromal tissue. The masked patch exhibits relatively compact cellular groups adjacent to pale stromal and vascular-like spaces, whereas the retained patch contains epithelial-like groups separated by more conspicuous fibrous stroma. Despite these local differences, both patches contain substantial cellular and architectural information. Their different masking outcomes illustrate that DARE does not simply remove cellular regions or exclude an entire tissue pattern. Instead, it selectively masks highly attended source-like regions while preserving other highly attended regions with overlapping morphological features.

A more direct morphological comparison is shown in Fig.~\ref{fig:masking_analysis}b. Both patches exhibit gland-like epithelial structures with pale or vacuolated cytoplasm. The masked patch contains more closely arranged epithelial structures, while the retained patch shows more open luminal spaces bordered by epithelial cells. Importantly, recognizable glandular morphology remains present in the retained patch. This example illustrates how selective masking can reduce reliance on a subset of highly attended regions without eliminating the broader morphological pattern available to the slide-level classifier.

Together, these examples highlight the complementary roles of attention and domain information: attention identifies regions emphasized by the classifier, while source-domain affinity guides which of these regions are masked. By selectively withholding highly attended source-like patches, DARE is designed to discourage reliance on source-associated appearance cues, which may include staining characteristics. Meanwhile, the retained regions provide alternative morphological evidence for slide-level aggregation. Within the teacher-student framework, this masked view encourages the student to maintain predictions consistent with the complete-slide teacher while drawing on the remaining tissue evidence.

\section{Domain Discrepancy and Adaptation Performance}
\label{sec:domain_discrepancy}

We investigate how the degree of domain discrepancy relates to the performance gains achieved by DARE. In addition to the cross-cohort experiments reported in the main text, we conduct supplementary cross-center experiments on CAMELYON17, which comprises data from five hospital centers. These experiments extend our evaluation to a multi-center setting and allow us to examine adaptation performance under relatively mild domain shifts.

To quantify domain discrepancy, we adopt the Fr\'echet-distance formulation underlying the Fr\'echet Inception Distance (FID)~\cite{heusel2017gans}, but compute it using CONCH embeddings instead of Inception features. We refer to this metric as the Fr\'echet Feature Distance (FFD). This choice allows us to measure domain mismatch in the representation space used by our downstream pathology models. Let $\boldsymbol{\mu}_s$ and $\boldsymbol{\mu}_t$ denote the empirical means of the source- and target-domain embeddings, and let $\boldsymbol{\Sigma}_s$ and $\boldsymbol{\Sigma}_t$ denote their covariance matrices. FFD is defined as
\begin{equation}
\mathrm{FFD}(s,t)
=
\left\|\boldsymbol{\mu}_s-\boldsymbol{\mu}_t\right\|_2^2
+
\mathrm{Tr}\left(
\boldsymbol{\Sigma}_s+\boldsymbol{\Sigma}_t
-
2\left(
\boldsymbol{\Sigma}_s^{1/2}
\boldsymbol{\Sigma}_t
\boldsymbol{\Sigma}_s^{1/2}
\right)^{1/2}
\right).
\end{equation}
A larger FFD indicates a greater discrepancy between the two feature distributions under this Gaussian approximation.

Table~\ref{tab:domain_discrepancy} summarizes FFD and the corresponding average AUCs of the source-only baseline and DARE. Across CAMELYON17 center pairs, FFD ranges from 0.0796 to 0.2469, with a mean of 0.1591. Under these relatively mild shifts, the source-only baseline already achieves an average AUC of 96.10\%, and DARE maintains comparable performance at 96.14\%. The strong baseline performance leaves limited room for further improvement.

Larger gains are observed in the cross-cohort settings with greater feature-space discrepancy. For TCGA-GBMLGG and IPD-Brain, an FFD of 0.3708 is accompanied by an average AUC improvement of 6.56 percentage points. For TCGA-NSCLC and CPTAC-NSCLC, the FFD reaches 0.7133, and DARE improves the average AUC by 20.16 percentage points. Together, these results show that DARE maintains strong average performance under mild center shifts and provides larger gains in the evaluated settings with more pronounced domain mismatch.

FFD measures overall feature-distribution mismatch, whereas adaptation performance also depends on how the shift affects task-relevant features and class boundaries. Thus, the observed relationship does not establish a universal discrepancy threshold for effective adaptation. The CAMELYON17 FFD range of 0.0796--0.2469 represents the smallest-discrepancy regime covered by our experiments.

\begin{table}[t]
\centering
\caption{Domain discrepancy and adaptation performance. FFD is computed using CONCH embeddings. AUCs are averaged over the evaluated transfer directions within each setting. Gain denotes the AUC improvement over the source-only baseline in percentage points (pp).}
\label{tab:domain_discrepancy}
\setlength{\tabcolsep}{4pt}
\resizebox{0.85\linewidth}{!}{%
\begin{tabular}{lcccc}
\toprule
Adaptation setting
& FFD
& \shortstack{Source-only\\AUC (\%)}
& \shortstack{DARE\\AUC (\%)}
& \shortstack{Gain\\(pp)} \\
\midrule
CAMELYON17 center pairs
& \shortstack{0.0796--0.2469\\(mean: 0.1591)}
& 96.10
& 96.14
& $+0.04$ \\
TCGA-GBMLGG $\leftrightarrow$ IPD-Brain
& 0.3708
& 77.11
& 83.67
& $+6.56$ \\
TCGA-NSCLC $\leftrightarrow$ CPTAC-NSCLC
& 0.7133
& 72.45
& 92.61
& $+20.16$ \\
\bottomrule
\end{tabular}%
}
\end{table}

\section{Computational Efficiency Analysis}
\label{sec:Computational Efficiency Analysis}

\begin{table}[htbp]
\centering
\caption{Training time comparison.}
\label{tab:training_time_detailed}
\begin{tabular}{l c c}
\toprule
Method & Total time (s)\\
\midrule
DANN-P   & 1183.66 \\
DANN         & 1202.09 \\
\midrule
DETA         & 6705.07 \\
RAM-MIL      & 2084.88 \\
\midrule
Ours         & 1721.52 \\
\bottomrule
\end{tabular}
\end{table}

The training time comparison is presented in Table~\ref{tab:training_time_detailed}.
Our method achieves a training time of 1721.52 seconds, which is significantly lower than that of RAM-MIL (2084.88 s) and DETA (6705.07 s), two representative UDA approaches designed for pathology.
Although our method is slower than the DANN-based baselines, those natural-image domain adaptation methods are less effective on histopathological data, as shown in the main experiments (Table~\ref{tab:main_resluts}).
Therefore, our method strikes a better balance between computational efficiency and domain adaptation performance in the pathology setting.

\section{Analysis of Mode Collapse}
\label{sec:mode_collapse_analysis}

As noted in the main results (Table~\ref{tab:main_resluts}), several compared methods produce identical accuracy (50.22\%) and F1-score (66.84\%) on the CPTAC \(\rightarrow\) TCGA (NSCLC) task. Mathematically, this specific metric combination indicates complete mode collapse: the adapted model loses all discriminative ability and degenerates into a trivial solution that predicts all target samples as the majority class.

We attribute this mode collapse to a deep inconsistency between prevailing alignment strategies and the distinct nature of whole slide image data. In WSIs, genuinely meaningful morphological signals, such as neoplastic patterns, are extremely sparse, whereas the representational space is crowded with domain-specific nuisance factors arising from differences in staining, scanner color response, and tissue handling. For generic unsupervised domain adaptation methods that rely on adversarial learning (e.g., DANN, CDAN), forcing a minimax game directly on such a heavily noisy latent space pushes the model to take the simplest route to deceive the domain classifier. Consequently, the model tends to shed not only domain-specific variations but also fragile diagnostic cues, driving the target representations into a tangled, semantically indiscernible state.

Notably, this vulnerability is not limited to generic methods but also exists in recently proposed pathology-specific UDA methods such as RAM-MIL and DETA. When RAM-MIL performs nearest-neighbor retrieval via optimal transport, the retrieved neighbors tend to be domain-similar rather than semantically similar instances. When DETA constructs graph structures in the feature space, it may establish spurious edges between patches that differ morphologically but share similar staining styles. These mechanisms, operating on input features already contaminated by strong domain bias, inadvertently amplify domain information and push the model further along the biased direction, eventually leading to the same degenerate solutions observed in other methods.

Our method avoids this pitfall. As shown in Fig.~\ref{fig:tsne_visualization}, we visualize the slide-level embeddings of three approaches: (a) the original DTFD backbone evaluated directly on the target domain, (b) DANN with slide-level adversarial training, and (c) our DARE. For DTFD, the two subtypes of target domain are completely inseparable. For DANN, a vague boundary remains poorly defined. In contrast, DARE produces a well-separated embedding space with a clear decision boundary between the two subtypes. This improvement stems from three synergistic designs. First, multi-granularity alignment jointly matches patch-level and slide-level feature distributions across domains, preventing the model from exploiting easy shortcuts that erase semantic information. Second, attention regularization encourages the model to focus on diagnostically relevant regions rather than domain-specific stains or artifacts. Third, the teacher-student pseudo-labeling module provides stable category-level supervision on the target domain, which guides the embedding space to preserve subtype discriminability even when domain shifts are severe. Together, these components enable DARE to recover meaningful structure without collapsing into a semantically indistinguishable state.

\section{Limitations and Future Work}
\label{sec:Limitations and Future Work}
\textbf{Limitations.}
First, although our approach is more efficient than existing pathology-specific UDA methods, its training speed is slower than methods that perform adversarial alignment at only one level. As shown in Table~\ref{tab:training_time_detailed}, the dual-level adversarial design and the teacher-student pseudo-label learning framework introduce extra computational overhead compared to single-level adaptation methods. 
Second, similar to other MIL-based WSI classification models, the performance of our method is influenced by the choice of the feature encoder. As reported in Table~\ref{tab:main_resluts} and Table~\ref{tab:encoder_robustness}, when switching from CONCH to the UNI2-h encoder, both our method and the compared baselines exhibit performance changes, with most methods achieving higher absolute scores under UNI2-h. While our method consistently achieves the best results across different encoders, the upper bound of overall performance is still tied to the quality of the underlying feature extractor.

\textbf{Future Work.}
A practical concern in real-world deployment is data privacy: source domain data may not always be accessible due to institutional regulations. In future work, we plan to extend our framework to source-free domain adaptation for WSIs, where only a pre-trained source model and unlabeled target data are available. This would help alleviate privacy concerns in clinical scenarios where direct sharing of source domain slides is restricted.

\section{Broader Impact}
\label{sec:impact}
Our work aims to improve the robustness of whole slide image classification under domain shifts commonly encountered across different medical centers. This line of research may help make pathology AI more accessible to hospitals that lack large annotated datasets, with the long-term potential to support more consistent diagnostic services.
We acknowledge that any AI-assisted diagnostic tool carries risks if deployed without proper validation. Our method, like other data-driven approaches, may still be susceptible to biases in the training data. We therefore advocate for rigorous external validation and close collaboration with clinical experts before real-world deployment.

\end{document}